\documentclass{llncs}

\usepackage[T1]{fontenc}
\usepackage{array}
\usepackage{graphicx}
\usepackage{xspace}
\usepackage{enumerate}
\usepackage{courier} 
\usepackage[table,xcdraw]{xcolor}
\usepackage{amsmath}
\usepackage{algorithm2e}

\usepackage{multirow} 

\usepackage{hyperref}
\usepackage{color}

\SetKwInput{KwInput}{Input}
\SetKwInput{KwOutput}{Output}

\newcommand{\owlvec}{\texttt{OWL2Vec*}\xspace}
\newcommand{\owlvecOA}{\texttt{OWL2Vec4OA}\xspace}
\newcommand{\ie}{\textit{i}.\textit{e}.,\xspace}
\newcommand{\eg}{\textit{e}.\textit{g}.,\xspace}

\newcommand{\mapping}[4]{\langle #1,\allowbreak #2,\allowbreak #3,\allowbreak #4 \rangle}

\begin{document}

%
\title{\owlvecOA: Tailoring Knowledge Graph Embeddings for Ontology Alignment}
\titlerunning{Tailoring Knowledge Graph Embeddings for Ontology Alignment}
%
\author{Sevinj Teymurova\inst{1} \and
Ernesto Jim\'enez-Ruiz\inst{1,2} \and Tillman Weyde\inst{1} \and \\ Jiaoyan Chen\inst{3}}
\institute{City St George's, University of London, UK \and University of Oslo, Norway \and University of Manchester, UK}
\maketitle              
\begin{abstract}
Ontology alignment is integral to achieving semantic interoperability as the number of available ontologies covering intersecting domains is increasing. This paper proposes \owlvecOA, an extension of the ontology embedding system \owlvec. While \owlvec has emerged as a powerful technique for ontology embedding, it currently lacks a mechanism to tailor the embedding to the ontology alignment task.
\owlvecOA incorporates 
edge confidence values from seed mappings to guide the random walk strategy. We present the theoretical foundations, implementation details, and experimental evaluation of our proposed extension, demonstrating its potential effectiveness for ontology alignment tasks.

\end{abstract}

\section{Introduction}


Knowledge graphs (and ontologies) are increasingly recognized as essential for successful AI implementations across various data science applications \cite{fensel2020knowledge}. 
Ontology alignment is a crucial task to enable 
semantic interoperability and enhance the application of knowledge graphs. The ontology alignment process involves finding and harmonizing semantic connections between different ontologies. While current methods have advanced ontology alignment considerably, there remain obstacles to their widespread implementation \cite{Harrow2019OntologyMF,DBLP:journals/ker/LiDFIJLP19}.

The ontology matching community has contributed to the evolution of ontology alignment systems for the last twenty years with the organization of the annual Ontology Alignment Evaluation Initiative (OAEI) \cite{oaei2022results,oaei2023}. In recent years there has been a shift from traditional systems \cite{euzenat2011ontology}, using lexical and structural techniques, to systems using machine learning
and (large) language models. 
Prominent examples include LogMap-ML \cite{chen2021augmenting}, BertMap \cite{bertmap2022}, DeepAlignment \cite{kolyvakis2018deepalignment}, VeeAlign \cite{iyer2020veealign}, SORBETMatcher \cite{gosselin2023sorbet} and OLaLa \cite{DBLP:conf/kcap/HertlingP23}.
The OAEI, with the new Bio-ML track \cite{DBLP:conf/semweb/HeCDJHH22,DBLP:conf/semweb/0008C0023}, has also evolved accordingly to attract and systematically evaluate such systems.

Knowledge Graph Embeddings (KGE) techniques~\cite{DBLP:journals/tkde/WangMWG17,DBLP:journals/tkdd/RossiBFMM21} aim at capturing, in a low-dimensional continuous vector space, the structure and semantics of the graph. 
These low-dimensional representations enable the application of machine learning algorithms to graph-structured data in downstream tasks such as node classification, link prediction, or knowledge graph alignment \cite{cai2018comprehensive}.
Traditional KGE techniques commonly rely on one of the following: \textit{(i)} geometric transformations, \textit{(ii)} matrix factorization methods, and \textit{(iii)} neural networks. 
Recent advancements in KGE have expanded to incorporate semantics beyond relational facts. These include encoding textual literals and integrating logical structures to capture richer semantic information within KG representations \cite{chen2024ontology}.
 
In this paper, we present \owlvecOA, an extension of the ontology embedding system \owlvec  \cite{chen2021owl2vec} tailored to the ontology alignment task. 
\owlvec projects a given ontology into a graph, randomly walks over the graph to generate sequences of entities, and runs the language model Word2Vec \cite{mikolov2013distributed} to generate embeddings of both entity URIs and words.
\owlvecOA, unlike \owlvec, relies on potentially incomplete or inaccurate ontology alignments to bridge a given set of input ontologies. When projecting the ontologies and performing the random walks to create entity sequences, the confidence value of these seed mapping are used to bias the random walks (\ie edges with higher confidence values will have higher chances to be visited). Hence, \owlvecOA allows for a tighter connection of the input ontologies given a set of seed mappings, while giving preference to edges with higher confidence.\footnote{Edges projected from ontology axioms are given the highest confidence.} 
\owlvecOA currently relies on the ontology matching systems LogMap \cite{cuenca2011logmap} and AML \cite{DBLP:conf/otm/FariaPSPCC13} to produce seed mappings.

Our experiments show that the embeddings computed by \owlvecOA are more suitable to the ontology alignment tasks than the original \owlvec vectors. \owlvecOA embeddings also lead to promising ranking results in the OAEI's Bio-ML track by simply comparing the computed vectors between the relevant source and target entities.

The rest of the paper is organised as follows. Section \ref{sec:prelim} introduces the necessary notions behind \owlvecOA. Section \ref{sec:related} presents relevant related work. \owlvecOA is described in detail in Section \ref{sec:methods}. Section \ref{sec:results} provide experimental results of \owlvecOA on the Bio-ML datasets. Finally, Section \ref{sec:conclusions} concludes the paper and discusses potential lines for future work.

\section{Preliminaries}
\label{sec:prelim}



\paragraph{Ontologies and knowledge graphs.}
\textit{Ontologies} serve as structured, clearly defined representations of collectively agreed-upon concepts and relationships within a specific field or area of interest~\cite{gruber1995toward}. Widely applied in information retrieval, data integration, and knowledge-based systems, ontologies facilitate semantic interoperability and reasoning across diverse applications. 
Knowledge graphs~\cite{hogan2021knowledge} have recently gained attention to represent entities and relationships within a graph-structured data model and have been very successful to improve search functionality, tailor user experiences, and inform strategic business choices \cite{fensel2020knowledge}. Nonetheless, from the Semantic Web point of view, in essence, ontologies and knowledge graphs can be seen as equivalent notions (\ie OWL ontologies provide a formalization of the RDF graph data model \cite{DBLP:journals/ws/GrauHMPPS08}, while knowledge graphs also imply the existence of such formalization). In this paper, we use knowledge graphs and ontologies interchangeably.

\begin{figure}[t!]
\centering
\includegraphics[width=0.99\textwidth]{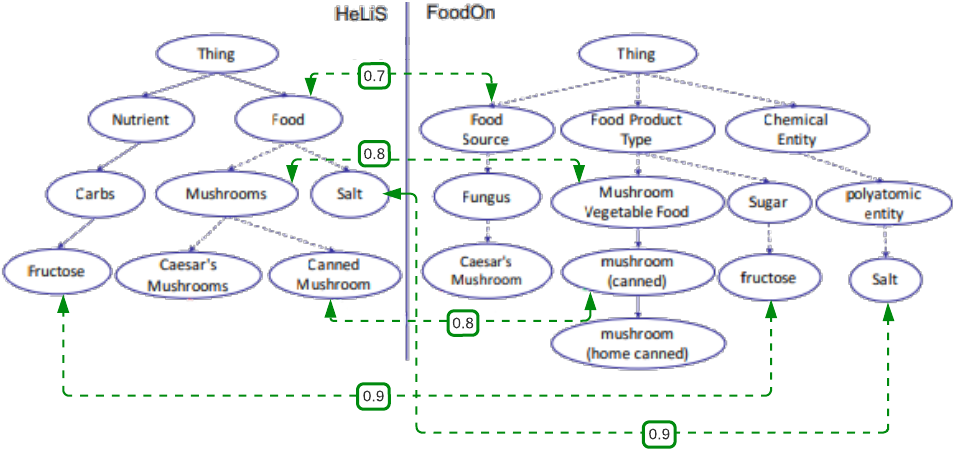}
\caption{Fragment of an alignment between HeLiS and FoodOn (adapted from~\cite{chen2021augmenting}). The green dash arrow denotes mappings with confidence values ranging from [0,1]. Blue arrows represent the inverse of the predicate \texttt{rdfs:subClassOf}.
} \label{fig:alignment}
\end{figure}

\paragraph{Ontology alignment}
is essential for data integration, semantic search, and cross-ontological reasoning. Ontology alignment can be defined as the process of identifying semantic relationships between elements (such as classes, attributes, and instances) of two or more ontologies. In this paper, we focus on atomic ontology matching where the goal is to establish equivalence or  subsumption
among atomic (\ie named) entities in the input ontologies \cite{euzenat2011ontology,otero2015ontology}. 
Mappings are typically represented as a 4-tuple $\mapping{e}{e'}{r}{c}$ where $e$ and $e'$ are entities from different ontologies; $r$ is a semantic relation (\eg equivalence or subsumption); and $c$ is a confidence value, usually, a real number within the interval $\left(0 \ldots 1\right]$.
For instance, Figure \ref{fig:alignment} shows a fragment of an alignment between the ontologies HeLiS~\cite{dragoni2018helis} and FoodOn~\cite{dooley2018foodon}.
The alignment indicates that the concept \texttt{HeLiS:Fructose} is similar to the concept \texttt{obo:FOODON\_03301305} (fructose) with a confidence value of 0.9. Confidence values are typically provided by ontology alignment systems and represent the degree of certainty associated with a correspondence between entities.
Alignment systems often employ sophisticated algorithms that consider lexical similarities, structural relationships, and external knowledge sources. Methods such as cross-referencing, semantic similarity measures, and machine learning techniques can be applied to establish the mappings.


\begin{figure}[t!]
    \centering
    \includegraphics[width=0.99\linewidth]{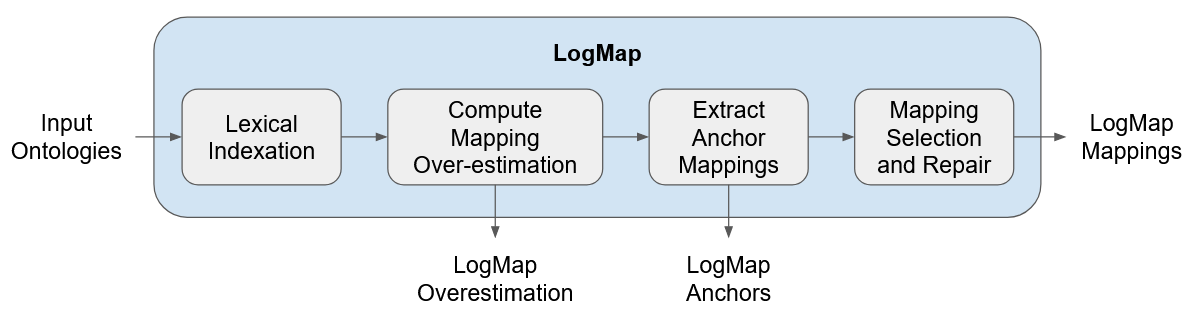}
    \caption{General architecture of LogMap.}
    \label{fig:logmap}
\end{figure}

\paragraph{LogMap} \cite{cuenca2011logmap,logmap2012} is an efficient ontology alignment tool for large-scale ontologies which employs lexical indexation, logic-based reasoning, and semantic similarity computation in a multi-stage process. LogMap has demonstrated effectiveness in various challenges and applications, especially in biomedicine.
As shown in Figure \ref{fig:logmap}, LogMap produces as output three different mappings sets: \textit{(i)} LogMap overestimation are a large set of candidate mappings aiming for high recall while representing a manageable subset of all possible mappings; \textit{(ii)} LogMap anchors are typicall a highly precise set of mappings; and \textit{(iii)} LogMap mappings are the final computed mappings aiming at a balanced Precision and Recall. These set of mappings will be used in our experiments.



\paragraph{Random walks} are key for embedding systems like RDF2Vec \cite{ristoski2016rdf2vec}, \owlvec and node2vec \cite{grover2016node2vec} to capture the structural and contextual information of a graph or network, so that the system can learn about the relationships between different nodes and their local neighborhoods. The PageRank algorithm \cite{lawrence1998pagerank} revolutionized web page ranking by employing a random walk model on the web's hyperlink structure. It simulates a ``random surfer'' traversing a graph of web pages, computing page importance based on the probability of the surfer landing on each page. This approach effectively captures the web's complex link topology to determine page significance, becoming a cornerstone in information retrieval and influencing various fields beyond web search. 
Random walks allow to process massive graphs without needing to consider all possible paths. Seminal works like DeepWalk \cite{perozzi2014deepwalk} and node2vec\cite{grover2016node2vec} utilize random walks to generate node sequences for training embeddings, effectively capturing network topology.  Wei \cite{wei2004towards} proposed an extension of the Metropolis-Hastings algorithm for sampling from large-scale networks, introducing strategies such as early rejection and biased sampling. RDF2Vec \cite{ristoski2016rdf2vec} introduced random walk-based embeddings for RDF knowledge graphs, later extended by Steenwinckel \cite{steenwinckel2023pyrdf2vec} with new walk extraction strategies.
Cochez et al. \cite{DBLP:conf/wims/CochezRPP17} also introduced biased walk strategies to RDF2Vec with twelve different edge weighting functions.


    \begin{figure}[t!]
        \centering
    \includegraphics[width=0.75\textwidth]{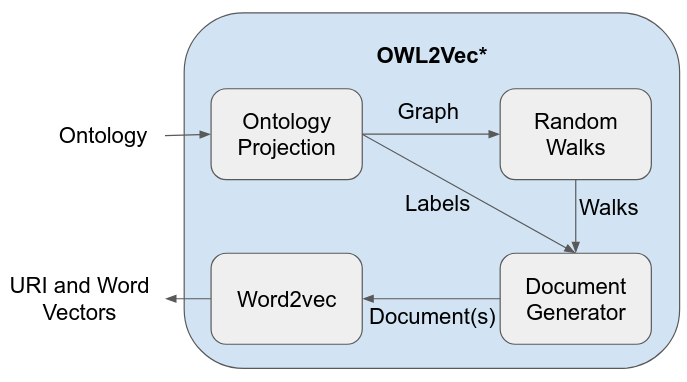}
        \caption{General architecture of \owlvec}
        \label{fig:owl2vec}
    \end{figure}

\paragraph{OWL2Vec.} Inspired by RDF2Vec \cite{ristoski2016rdf2vec}, \owlvec~\cite{chen2021owl2vec} was designed following similar principles but adapted to create embeddings for OWL ontologies and to take into account the lexical information of the ontologies (\eg literals in the form of labels and synonyms).
Figure \ref{fig:owl2vec} depicts the architecture of \owlvec for generating ontology embeddings. \owlvec projects the input ontology into a graph and generates entity sequences via random walks over the ontology graph, then it generates different types of documents by substituting none, some, or all entity URIs by its lexical representation. Finally, the word embedding model Word2vec \cite{mikolov2013distributed} is applied over the generated documents to compute embeddings for both URIs and words. Note that URIs are unique and thus their embeddings are contextual.
\owlvec has shown to outperform other approaches in intra-ontology subsumption and class membership prediction when both structural and lexical information was critical. \owlvec, however, focuses on creating embedding for a single ontology. Although a set of ontologies can also be given as input, their graph representation will not be connected and thus the random walks will not generate sequences involving elements from different ontologies.

\section{Related Work}
\label{sec:related}


The ongoing research in the ontology matching community evidences the need for more sophisticated techniques, as shown in the annual OAEI campaign~\cite{oaei2023} new methodologies and systems are developed to address this challenging problem in the Semantic Web. Otero-Cerdeira et al. \cite{otero2015ontology} provides a comprehensive survey of ontology matching techniques. 

These diverse approaches demonstrate the complexity of ontology alignment and the variety of techniques employed \cite{DBLP:journals/biomedsem/FariaPMMCC18}.
Nonetheless, ontology matching tools can now process more efficiently even the most complex ontologies, including those with hundreds of thousands of classes, encompassing billions of possible connections. 
Recent advancements in ontology matching have incorporated machine learning, including embedding-based techniques and (large) language models, showing promising results in improving alignment accuracy (\eg \cite{chen2021augmenting}, \cite{bertmap2022}). These approaches leverage vector representations of ontological elements to capture semantic relationships more effectively than traditional methods. By utilizing language models or domain-specific embedding algorithms, these techniques can identify nuanced similarities between concepts, potentially leading to more accurate and comprehensive ontology alignments. This new generation of systems can broadly be categorized into three categories.
\textit{(i) Direct embedding comparison:} Methods like ERSOM \cite{xiang2015ersom} and DeepAlignment \cite{kolyvakis2018deepalignment} calculate distances between concept embeddings directly.
\textit{(ii) Supervised mapping classifiers:} VeeAlign \cite{iyer2020veealign}, MEDTO \cite{hao2021medto}, LogMap-ML \cite{chen2021augmenting}, 
and SORBET \cite{gosselin2023sorbet} train classifiers using concept embeddings as input. This approach adds a layer of learning specific to the OM task but still relies on independent embeddings for the input ontologies.
\textit{(iii) Based on language models:} Recent methods such as BERTMap \cite{bertmap2022}, BERTSubs \cite{chen2023contextual} and OLaLa \cite{DBLP:conf/kcap/HertlingP23} rely on language models to implement task-specific models. For instance, BERTMap fine-tunes pre-trained language models (PLMs) using synonyms from the ontologies, while BERTSubs focuses on subsumption mapping prediction using context-based information that is transformed into text by templates.

Although systems based on language models have shown impressive results, they rely on pre-trained or large language models adding an important complexity layer for large matching tasks. Our approach \owlvecOA leverages a simpler language model like Word2Vec to create tailored embeddings for the ontology matching task. 
Although Word2Vec does not create contextual embeddings (\ie same string with different meanings will get the same embedding), the sequences that \owlvecOA creates include \textit{things} (\ie entity URIs) in addition to strings, leading to contextual embeddings for the ontology entities as URIs are unique. The embeddings are tailored to the matching task as \owlvecOA bridges the input ontologies with seed mappings computed by a (traditional) alignment system like LogMap \cite{cuenca2011logmap} and AML \cite{DBLP:conf/otm/FariaPSPCC13} before performing the random walks to generate the URI and word sequences. Hence, unlike other approaches like LogMap-ML, the computed embeddings for both ontologies are in the same vector space and tightly related via the seed mappings.

Our evaluation uses the datasets of the OAEI's Bio-ML track \cite{DBLP:conf/semweb/HeCDJHH22}, specialized benchmarks for evaluating machine learning-based OM systems. Currently, our experiments focus on direct embedding comparison, as the main purpose of this exercise was to evaluate the quality of the embeddings without any additional layer. The reported results are promising. In the near future we also plan to train a model with the computed embeddings similarly to LogMap-ML and the approach presented by Hao et al. \cite{DBLP:journals/ws/HaoMXLQF23}.

\begin{figure}[t!]
    \centering
\includegraphics[width=0.99\textwidth]{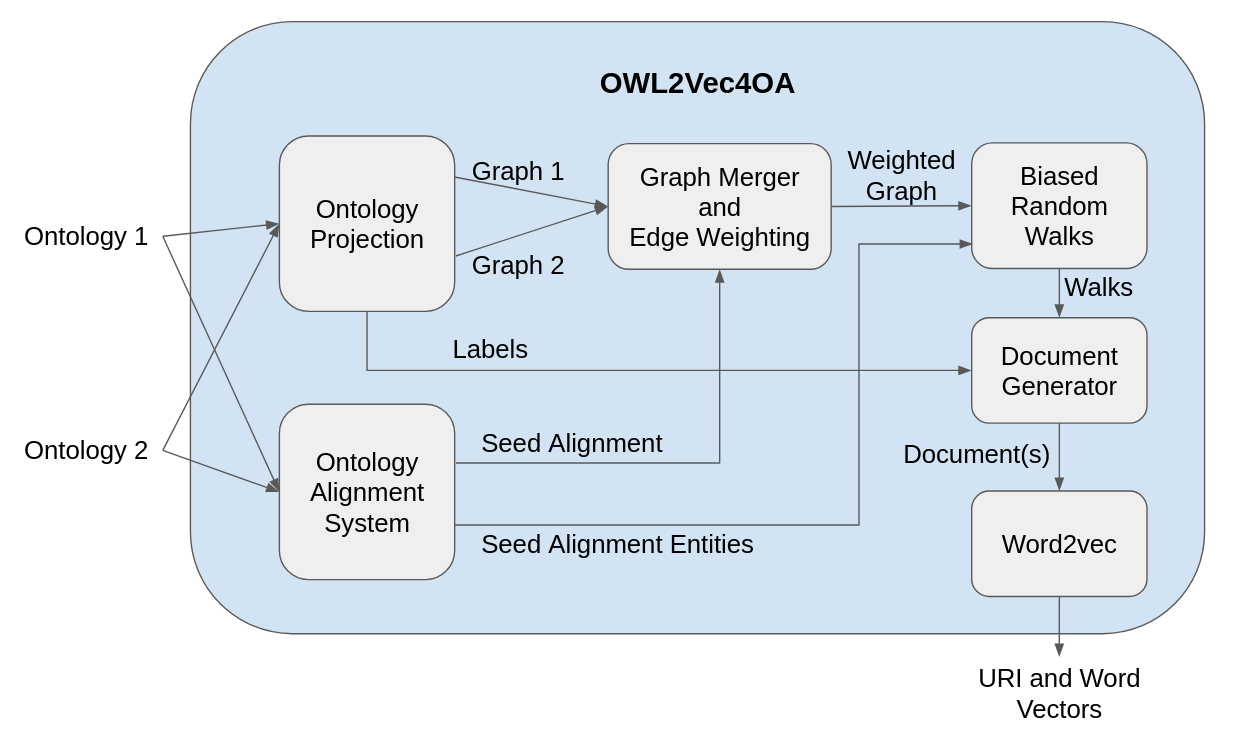}
    \caption{General architecture of \owlvecOA}
    \label{fig:owl2vec4oa}
\end{figure}

\section{Ontology embeddings with \owlvecOA}
\label{sec:methods}

\owlvecOA extends the \owlvec system with a mechanism to tailor the embeddings to the ontology alignment task involving two or more input ontologies.
The main steps of our \owlvecOA are depicted in Figure \ref{fig:owl2vec4oa},\footnote{Source codes of \owlvecOA are available here: \url{https://github.com/Sevinjt/OWL2Vec4OA}} 
and summarised as follows.

\paragraph{Ontology projection.} We use the same ontology projection rules as in \linebreak \owlvec~\cite{chen2021owl2vec}. The projection rules transform one or more ontology axioms into RDF triples (\ie labeled edges in the projected graph). Some axioms such as class subsumption (\eg \texttt{obo:CHEBI\_28757} (fructose) \texttt{\textbf{rdfs:subClassOf}} \texttt{obo:FOODON\_03420108} (sugar)) and annotations (\eg  \texttt{obo:FOODON\_03420108} \texttt{\textbf{rdfs:label}}  \texttt{"sugar"}) have a one-to-one triple transformation; while more complex axioms require the application of projection rules. For example, 
the axiom \texttt{obo:FOODON\_03301391} (mushroom (canned)) \texttt{\textbf{rdfs:subClassOf}} \linebreak \texttt{RO\_0001000} (derives from)
\texttt{\textbf{some} FOODON\_03411261} (fungus)
is transformed into the triple  $\langle$~\texttt{obo:FOODON\_03301391}, \texttt{RO\_0001000}, \texttt{FOODON\_03411261}~$\rangle$.  A directed labeled graph for each of the input ontologies is returned as the output of the projection.

\paragraph{Ontology alignment.} 
\owlvecOA currently relies on the traditional ontology matching systems LogMap \cite{cuenca2011logmap} and AML \cite{DBLP:conf/otm/FariaPSPCC13} to produce seed mappings. 
For example, Figure \ref{fig:alignment} shows a subset of plausible mappings computed by an alignment system.
These seed mappings are used to bridge the ontology graphs and thus enabling the execution of random walks over entities from different ontologies. 
Note that, seed mappings do not need to be accurate nor complete and their confidence values will be used to bias the random walks. As shown in Figure \ref{fig:logmap}, LogMap produces three sets of mappings of different quality that will be used as seed: \textit{(i)} an overestimation of potential mappings (LogMap$_{over}$), \textit{(ii)} highly precise mappings or anchors (LogMap$_{anch}$), and \textit{(iii)} the regular output mappings (LogMap$_{out}$). In addition, LogMap$_{out}$ mappings are combined with AML mappings in our experiments (\ie LogMap$_{out}$ $\cup$ AML, and LogMap$_{out}$ $\cap$~AML).

\paragraph{Graph merger and Edge Weighting.} Unlike 
our predecessor
\owlvec, \linebreak \owlvecOA builds a single graph taking as input the graph projections of the ontologies to be aligned and a set of seed mappings.
Mappings (\ie $\mapping{e}{e'}{r}{c}$) have a direct graph representation as triples, for example, in Figure \ref{fig:alignment} the following  mapping ($r=$ equivalence) was identified 
$\langle$ \texttt{HeLiS:Fructose}, \linebreak \texttt{\textbf{owl:equivalentClass}}, \texttt{obo:FOODON\_03301305} (fructose) $\rangle$ with confidence $c$ = 0.9. 
\owlvecOA assigns a weight to each edge or link as follows: \textit{(i)} $1.0$ if the edge was derived from ontology axioms; and \textit{(ii)}~$c$~if the edge was derived from a mapping. The output is a labeled weighted graph $G=(V, E, U, W)$, where $E$ is the set of edges built from the projected RDF triples, $V$ is the set of vertices composed by the subjects and objects in these triples, $U$ is the the set of URIs associated to the vertices and edges, and $W$ is the set of weights associated to the edges. The function $weight(G, l)$ returns the weight for a given edge $l$, while the function $uri(G, e)$ returns the URI of a given entity $e$.

\begin{algorithm}[t]
\SetAlgoLined
\KwInput{Weighted graph $G =(V, E, U, W)$; seed entities $S$; walk depth $wd$\; iterations $iter$ }
\KwOutput{Walks or entity sequences $W$}
W = \{\};

\For{k \textbf{in} range($iter$)}{
\tcp{iterates over the seed entities}
\For{e \text{\textbf{in}} S}{
$current\_walk = [~]$\;
Append $uri(G, e)$ to $current\_walk$\;
$current\_size = 1$\;
$focus = e$\;
\While{$current\_size < wd$}{
Extract set of outer edges $E_{focus}$ from $focus$ vertex\;
\If{$|E_{focus}|=0$}{
\textbf{break}\;
}
\tcp{According to the probabilities of the edges in $E_{focus}$ as in Equation 1}
Randomly select an outer edge $l$ such that $l=(focus, v) \in E_{focus}$\;
Append $uri(G, l)$ to $current\_walk$; \tcp{URI of the link}
Append $uri(G, v)$ to $current\_walk$; \tcp{URI of the vertex} 
$current\_size++$\;
$focus=v$\;
}
Add $current\_walk$ to $W$
}
}
 \Return $W$\tcp*{Set of walks/sequences.}
 \vspace{0.25cm}
 \caption{
Biased Random Walks Algorithm
 }\label{alg:biasalgo}
\end{algorithm}

\paragraph{Biased random walks.} \owlvecOA, inspired by Cochez et al. \cite{DBLP:conf/wims/CochezRPP17}, implements the biased random walker summarised in Algorithm \ref{alg:biasalgo}. The algorithm takes as input a weighted labeled graph $G$ and a set of seed entities $S$ and performs (biased) random walks of depth $wd$ starting from each of the seed entities. It optionally iterates over the seed entities more than once to allow for different walks for the same seed entity. In our setting, the seed entities represent the entities involved in the seed mappings computed in the alignment step. This way the walks are tailored to the alignment task, without the need of exploring the whole input ontologies. The bias in the random walk takes into account the weight assigned to each of the edges or links ($l$) in the graph $G$ to assign a probability to each of the potential paths.
%
Given $E_u=\{l_i = (u, v_i) \text{~with~} i=1..n\}$ and $l_i$ an outer edge for $u$, the probability for each edge is computed as in Equation \ref{eq:probability}.
\begin{align}
    Pr(l_j=(u, v_j)) = \frac{weight(G, l_j)}{\sum_{i=1}^n weight(G, l_i)}
    \label{eq:probability}
\end{align}

For example, a walk of depth 3 starting from the seed entity \texttt{HeLiS:Fructose} (\ie an entity appearing in a mapping) could include the following sequence of URIs:
\texttt{HeLiS:Fructose},
\texttt{owl:equivalentClass},
\texttt{obo:FOODON\_03301305} (fructose),
\texttt{rdfs:subClassOf}, 
\texttt{obo:FOODON\_03420108} (sugar).

\paragraph{Document generator and Word2Vec embeddings.} \owlvecOA, as in \owlvec, creates three types of documents from the generated walks $W$ in the previous step: \textit{(i)} structure document, \textit{(ii)} lexical document, and \textit{(iii)} combined document. The structure document is a direct representation of the walks as the sentences are composed by entity URIs. The lexical document replaces every URI occurrence in the walks by the respective lexical representation of the entity (\ie the occurrence of \texttt{obo:FOODON\_03420108} is replaced by its lexical representation ``sugar'', typically provided in the ontology via an \texttt{rdfs:label} annotation). Finally, the combined document randomly replaces in each walk some of the entity URI occurrences by its associated label.
The three documents are merged and used to train a Word2Vec model with the skip-gram architecture. The trained Word2Vec model produces embeddings for both URI and word occurrences in the merged document.
As introduced in Section \ref{sec:prelim}, the URI embeddings can be seen as contextual embeddings as URIs are unique.


\section{Evaluation}
\label{sec:results}

We have performed a preliminary evaluation of the suitability of the embeddings computed by \owlvecOA in ontology alignment tasks. Particularly we have used the datasets provided by the OAEI's 2023 Bio-ML track\footnote{Bio-ML Challenge \cite{DBLP:conf/semweb/HeCDJHH22}: \url{https://krr-oxford.github.io/OAEI-Bio-ML/}. Bio-ML 2023 Datasets: \url{https://doi.org/10.5281/zenodo.8193375}}. The Bio-ML track included several tasks (\eg OMIM-ORDO, NCIT-DOID, SNOMED-NCIT-Pharm and SNOMED-NCI-Neoplas), involving biomedical ontologies with tens of thousands of classes, and reference alignments based on Mondo \cite{Vasilevsky2022.04.13.22273750} and UMLS \cite{DBLP:journals/nar/Bodenreider04}.

Bio-ML presents two evaluation settings: global matching and local ranking. Global matching is evaluated with the traditional measures Precision and Recall, comparing a set of system-computed mappings with the reference set of mappings; while local matching evaluates the capacity of a system to rank a correct mapping given a pool of potential candidates. Bio-ML uses Mean Reciprocal Rank (MRR) and Hits@K (\ie cases where the correct mapping was ranked within the top-k) in the local matching setting.

\paragraph{Scoring function and settings.}
We have applied \owlvecOA embeddings into the local matching tasks of Bio-ML. Mappings are scored and ranked according to the cosine similarity of the computed URI embeddings for the entities in the mapping. 
We have computed \owlvecOA embeddings for different walk depths and iterations. We fixed the Word2Vec hyperparameters --- the number of epochs and embedding dimension to 70 and 100, respectively. The experiments were conducted on a High-Performance Computing cluster with access to up to 48 CPUs, using the Slurm workload manager to ensure efficient resource allocation and job scheduling.
Generated resources are available in Zenodo \cite{owl2vec4ow_zenodo}.



\paragraph{Impact of seed mappings and number of iterations.} Table \ref{tab:results1} shows the results over the OMIM-ORDO for different sets of seed mappings as introduced in Section \ref{sec:methods}. We also used as seed the mappings provided as training and validation in Bio-ML, as shown in the first row, using training-validation mappings in isolation did not lead to promising results given their reduced size. The set of seed mappings leading to the best results was the union of LogMap$_{out}$ and AML mappings, with LogMap$_{out}$ $\cap$~AML and LogMap$_{over}$ leading to similar results. The results also show that \owlvecOA is also able to handle noisy set of seed mappings like LogMap$_{over}$.
The impact of additional iterations over the seed entities did not lead to a significantly increased performance.

\setlength{\tabcolsep}{5.5pt}
\begin{table}[t]
\centering
\caption{OMIM-ORDO task with Walk depth 3, Walker iteration:  iter=1 / iter=5.
}\label{tab:results1}
\begin{tabular}{|l|c|c|c|c|c|c|}
\hline
                                 \textbf{Seed Mappings}   & \textbf{Hits@1} & \textbf{Hits@5} & \textbf{Hits@10} & \textbf{MRR}    \\ \hline\hline
Train-Validation                           & 0.01 / 0.01   & 0.02 / 0.02  & 0.05 / 0.05    & 0.04 / 0.04 \\ \hline\hline
LogMap$_{over}$    & 0.27 / 0.28   & 0.51 / 0.53   & \textbf{0.60 / 0.62}   & 0.38 / 0.40  \\ \hline
LogMap$_{anch}$ & 0.11 / 0.26  & 0.28 / 0.41   & 0.36 / 0.46    & 0.20 / 0.33   \\ \hline
LogMap$_{out}$            & 0.26 / 0.27   & 0.41 / 0.45   & 0.47 / 0.53    & 0.33 / 0.36  \\ \hline
LogMap$_{out}$ $\cup$ AML       & 0.30 / 0.31  & \textbf{0.54 / 0.54}  & \textbf{0.61 / 0.61}   & \textbf{0.41 / 0.41} \\ \hline
LogMap$_{out}$ $\cap$~AML & \textbf{0.31 / 0.34}  & 0.49 / 0.50  & 0.54 / 0.54    & 0.40 / 0.41  \\ \hline
\end{tabular}
\end{table}

\setlength{\tabcolsep}{1.5pt}
\begin{table}[t]
\caption{Results of \owlvecOA and \owlvec over four Bio-ML tasks, with different walk depths (wd).}
\label{tab:combined_results}
\resizebox{\textwidth}{!}{%
\centering
\begin{footnotesize}
\begin{tabular}{|l|l|c|c|c|c|c|c|c|}
\hline
\multicolumn{1}{|c|}{\textbf{Task}} & \multicolumn{1}{c|}{\textbf{System}} & \textbf{~wd~} & \textbf{MRR} & \textbf{Hits@1} & \textbf{Hits@5} & \textbf{Hits@10} & \textbf{Hits@20} & \textbf{Hits@30} \\
\hline\hline
\multirow{6}{*}{OMIM-ORDO} & \multirow{3}{*}{\owlvec} & 
      2 & 0.074 & 0.018 & 0.091 & 0.178 & 0.332 & 0.393 \\
 &  & 3 & 0.073 & 0.018 & 0.090 & 0.170 & 0.318 & 0.381  \\
 &  & 4 & 0.071 & 0.019 & 0.078 & 0.320 & 0.321 & 0.387 \\
\cline{2-9}
 & \multirow{3}{*}{\owlvecOA} & 
      2 & \textbf{0.586} & \textbf{0.533} & \textbf{0.637} & \textbf{0.657} & \textbf{0.672} & \textbf{0.693}  \\
 &  & 3 & 0.402 & 0.306 & 0.512 & 0.587 &  0.650 & 0.685 \\
 &  & 4 & 0.215 & 0.132 & 0.281 & 0.359 &  0.446 & 0.532\\
\hline\hline
\multirow{6}{*}{NCIT-DOID} & \multirow{3}{*}{\owlvec} & 
      2 & 0.218 & 0.110 & 0.306 & 0.448 & 0.631  & 0.746 \\
 &  & 3 & 0.175 & 0.074 & 0.251 & 0.377 & 0.561  & 0.690 \\
 &  & 4 & 0.105 & 0.035 & 0.121 & 0.225 & 0.409  & 0.541 \\
\cline{2-9}
 & \multirow{3}{*}{\owlvecOA} & 
      2 & 0.195 & 0.064 & 0.310 & 0.508 & 0.709 & 0.812 \\
 &  & 3 & 0.358 & 0.181 & 0.573 & 0.741 & 0.872 & 0.924 \\
 &  & 4 & \textbf{0.609} & \textbf{0.442} & \textbf{0.840} & \textbf{0.928} & \textbf{0.970} & \textbf{0.984} \\
\hline\hline
\multirow{6}{*}{SNOMED-NCIT-N} & \multirow{3}{*}{\owlvec} & 
      2 & 0.063 & 0.014 & 0.075 & 0.134 & 0.231 & 0.309 \\
 &  & 3 & 0.068 & 0.017 & 0.079 & 0.142 & 0.238 & 0.308 \\
 &  & 4 & 0.055 & 0.011 & 0.052 & 0.114 & 0.218 & 0.305 \\
\cline{2-9}
 & \multirow{3}{*}{\owlvecOA} & 
      2 & 0.648 & 0.543 & 0.767 & 0.831 & 0.888  & 0.904 \\
 &  & 3 & 0.605 & 0.484 & 0.746 & 0.813 & 0.872 & 0.899 \\
 &  & 4 & \textbf{0.805} & \textbf{0.747} & \textbf{0.872} & \textbf{0.888} & \textbf{0.902} & \textbf{0.910} \\
\hline\hline
\multirow{6}{*}{SNOMED-NCIT-P} & \multirow{3}{*}{\owlvec} & 
      2 & 0.079 & 0.018 & 0.094 & 0.184 & 0.302  & 0.675 \\
 &  & 3 & 0.078 & 0.018 & 0.092 & 0.181 & 0.292  & 0.667 \\
 &  & 4 & 0.055 & 0.011 & 0.052 & 0.114 & 0.218 & 0.305 \\
\cline{2-9}
 & \multirow{3}{*}{\owlvecOA} & 2 & \textbf{0.436} & \textbf{0.342} & \textbf{0.534} & \textbf{0.583} & \textbf{0.609}  & \textbf{0.967} \\
 &  & 3 & 0.311 & 0.190 & 0.435 & 0.502 & 0.558  & 0.933 \\
 &  & 4 & 0.291 & 0.204 & 0.355 & 0.434 & 0.521  & 0.944 \\
\hline
\end{tabular}%
\end{footnotesize}
} 
\end{table}

\paragraph{Comparison with \owlvec.} Following the results in Table \ref{tab:results1}, we set LogMap$_{out}$ $\cup$ AML as the seed mappings and the number of iterations to 1 in the subsequent experiments. We experimented with walk depths ranging from 2 to 4.
We compared the performance of the embeddings computed with \owlvecOA with those computed with the original \owlvec version (using its multi-ontology setting). As expected, Table \ref{tab:combined_results} shows that the ranking with \owlvecOA embeddings considerably outperforms the ranking with \owlvec embeddings in all tasks and for all evaluated walk depths, indicating that the \owlvecOA embeddings are more suitable for ontology alignment tasks.
The best results are obtained for the task SNOMED-NCIT-Neoplas with walk depth $4$ where Hits@1 reach more than 80\% of the cases. In other tasks, the results are also promising indicating that the embeddings computed by \owlvecOA capture relevant features of the original entities that could be exploited by a subsequent machine learning model.

\paragraph{Impact of the walk depth.}
Increasing the walk depths has a positive impact in the tasks NCIT-DOID and SNOMED-NCIT-Neoplas; while for OMIM-ORDO and SNOMED-NCIT-Pharm longer paths seem to add noise to the embeddings. This is inline with the results obtained in the original \owlvec paper \cite{chen2021owl2vec} where longer paths did not seem to lead to better results.
It is worth mentioning that longer paths also increase the computation times.
In the near future, we plan to perform an extended evaluation to better understand the impact of longer walks on different ontologies and matching tasks.

\section{Conclusions and Future Work}
\label{sec:conclusions}

We have presented \owlvecOA, an extension of the ontology embedding system \owlvec  \cite{chen2021owl2vec}. \owlvecOA has been tailored to the ontology alignment task by using a preliminary set of ontology alignments, possibly incomplete or inaccurate, to bridge a given set of input ontologies. 
These seed mappings and their confidence are key when performing biased random walks to create sequences of entities from both ontologies. 
The results section shows promising results where the \owlvecOA embeddings lead to much better-ranking results than those computed by \owlvec.

Currently, our experiments rely on direct embedding comparison which leads to good similarity results. However, predicting equivalent or subsumption mappings is a more complex task. In the near future, we aim at training machine learning models to better benefit from the features of the \owlvecOA embeddings for an ontology alignment task. Prominent examples in the literature are LogMap-ML \cite{chen2021augmenting}, which successfully applied a Siamese Neural Network; and Hao et al. \cite{DBLP:journals/ws/HaoMXLQF23}, which explored the use of Graph Neural Networks (GNN). These approaches, however, created embeddings that were independent for each input ontology, unlike those computed by \owlvecOA.

In addition, we also plan to conduct additional experiments to better understand the impact of the walk depth with different strategies to create entity sequences (\ie focusing on concepts and/or avoiding OWL constructs). Entity embedding can also be constructed using the word embedding associated to their labels, which may bring additional features with respect to the URI embeddings. Finally, once we have an end-to-end ontology alignment system in place, we aim to participate in the OAEI campaign and perform an extensive comparison with the state-of-the-art.

\section*{Acknowledgements}
This research is funded by the Ministry of Education and Science of Azerbaijan Republic with support from City St George’s, University of London. This work has also been partially supported by the Academy of Medical Sciences Network Grant (Neurosymbolic AI for Medicine, NGR1\textbackslash1857), the project "XAI4SOC: Explainable Artificial Intelligence for Healthy Aging and Social Wellbeing" funded by the Agencia Estatal de Investigación (AEI), the Spanish Ministry of Science, Innovation and Universities and the European Social Funds (PID2021-123152OB-C22), the EPSRC project OntoEm (EP/Y017706/1), and the EU Projects: RE4DY (101058384, HORIZON-CL4-2021), Plooto (101092008, HORIZON-CL4-2022), SM4RTENANCE (101123490, DIGITAL-2022), and \linebreak Tec4MaasEs (101138517, HORIZON-CL4-2023).

%
%
%
\bibliography{bib}
\bibliographystyle{splncs04}

\end{document}